# Representation of a Sentence using a Polar Fuzzy Neutrosophic Semantic Net

Sachin Lakra
Research Scholar
Computer Science & Engineering
K. L. University Vaddeswaram,
Guntur, AP, India

T. V. Prasad
Former Dean of Computing
Sciences, Visvodaya Technical
Academy, Kavali, AP,
India.

G. Ramakrishna
Professor Computer Science &
Engineering K. L. University
Vaddeswaram, Guntur, AP,
India

***Abstract*—A semantic net can be used to represent a sentence. A sentence in a language contains semantics which are polar in nature, that is, semantics which are positive, neutral and negative. Neutrosophy is a relatively new field of science which can be used to mathematically represent triads of concepts. These triads include truth, indeterminacy and falsehood, and so also positivity, neutrality and negativity. Thus a conventional semantic net has been extended in this paper using neutrosophy into a Polar Fuzzy Neutrosophic Semantic Net. A Polar Fuzzy Neutrosophic Semantic Net has been implemented in MATLAB and has been used to illustrate a polar sentence in English language. The paper demonstrates a method for the representation of polarity in a computer's memory. Thus, polar concepts can be applied to imbibe a machine such as a robot, with emotions, making machine emotion representation possible.**

## I. INTRODUCTION

Representation of polarity, that is, positivity, neutrality and negativity, of a sentence in a natural language, has been a long-standing problem in Natural Language Processing. Knowledge representation using various techniques including frames, conceptual dependency and semantic nets [1] were proposed 7-8 decades ago since the advent of AI. Numerous accounts of artificially intelligent machines which are incapable of emotion representation exist in the form of science fiction literature. Machines are considered to be incapable of possessing emotions in the same way as human beings.

Semantic nets were first proposed by Charles S. Peirce in the year 1909 [2]. Semantic nets were first invented for computers by Richard H. Richens of the Cambridge Language Research Unit in 1956 [3]. An extension of semantic nets to include inexactitude and imprecision was made by the development of Fuzzy Semantic Nets [4]. Fuzzy Cognitive Maps and Neutrosophic Cognitive Maps were introduced in 2003 by Kandasamy and Smarandache [5]. These maps introduced the notion of causality in a network structure to represent concepts and their interdependence. However, none of these attempts were able to incorporate polarity.

This paper extends traditional semantic nets into polar fuzzy neutrosophic semantic nets (PFNSN). The term "polarity" has been introduced to distinguish the triad of the concepts of positivity, neutrality and negativity from the triad of concepts of truth, indeterminacy and falsehood. Earlier, both these triads were categorized under the term neutrosophy. A PFNSN has been implemented in MATLAB to represent an English sentence which specifically depicts polar concepts. Applications of a PFNSN can be found in machine emotion representation and intelligent response generation.

## II. NEUTROSOPHY

The name neutrosophy is derived from Latin "neuter" meaning neutral and Greek "sophia" meaning skill/wisdom. Neutrosophy is a branch of philosophy, introduced by Florentin Smarandache in 1980, which studies the origin, nature, and scope of neutralities, as well as their interactions with different ideational spectra [6].

Florentin Smarandache had generalized the fuzzy logic, and introduced two new concepts [7]:

- "neutrosophy" – study of neutralities as an extension of dialectics;
- and its derivative "neutrosophic", such as "neutrosophic logic", "neutrosophic set", "neutrosophic probability", and "neutrosophic statistics" and thus opened new ways of research in four fields: philosophy, logics, set theory, and probability/statistics.

Neutrosophy considers a proposition, theory, event, concept, or entity, "A" in relation to its opposite, "Anti-A" (and that which is also not A, "Non-A"), and that which is neither "A" nor "Anti-A", denoted by "Neut-A".

Neutrosophy is the basis of neutrosophic logic, neutrosophic probability, neutrosophic sets, and neutrosophic statistics.

Definition 1: Main Principle of Neutrosophy [6]

Between an idea <A> and its opposite <Anti-A>, there is a continuum-power spectrum of neutralities <Neut-A>.

Definition 2: Fundamental Thesis of Neutrosophy [6]

Any idea <A> is *t*% true, *i*% indeterminate, and *f*% false, where $t, i, f \in \, ]^-0, 1^+[$.

Here, $^-0 = 0 - \varepsilon$ and $1^+ = 1 + \varepsilon$, where $\varepsilon$ is an infinitesimal value.Definition 3:Neutrosophic Components [10].

Let *T, I, F* be standard or non-standard real subsets of $]^-0, 1^+[$, where the sets *T, I, F* are not necessarily intervals, but may be any real sub-unitary subsets: discrete or continuous; single-element, finite, or (countable or uncountable) infinite; union or intersection of various subsets; etc.

*T, I, F*, called neutrosophic components, represent the truth value, indeterminacy value, and falsehood value, respectively, referring to neutrosophy, neutrosophic logic, neutrosophic set, neutrosophic probability and neutrosophic statistics[10].

This representation is closer to the reasoning performed by the human mind. It characterizes/catches the imprecision of knowledge or linguistic inexactitude perceived by various observers (reason why *T, I, F* are subsets - not necessarily single-elements), uncertainty due to incomplete knowledge or acquisition errors or stochasticity (reason why the subset *I* exists), and vagueness due to lack of clear contours or limits (reason why *T, I, F* are subsets and *I* exists) [10].

One has to specify the superior (x_sup) and inferior (x_inf) limits of the subsets because in many problems the necessity to compute them arises.

## III. MATHEMATICAL DEFINITION OF A POLAR FUZZY NEUTROSOPHIC SEMANTIC NET

A semantic net has a graph as its core mathematical structure. Consequently, since a neutrosophic semantic net is an extension of a conventional semantic net, its core mathematical structure is a neutrosophic graph.

### A. *Neutrosophic Graph*

Definition 4: Graph

A graph is an ordered pair $G = (V, E)$ comprising set $V$ of vertices or nodes together with a set E of edges or lines, which are 2-element subsets of $V$ (i.e., an edge is related with two vertices, and the relation is represented as an unordered pair of the vertices with respect to the particular edge) [8, 9].

Definition 5: Neutrosophic Set [10]

A Neutrosophic Set is a set such that an element belongs to the set with a neutrosophic probability, i.e. *t*% is true that the element is in the set, *f*% false, and *i*% indeterminate.

Definition 6: Neutrosophic Indeterminacy

Indeterminacy is defined as the state of being defined in an inexact manner. Indeterminacy is represented in neutrosophy by a set I of values referring to the degree of inexactitude and is termed as Neutrosophic Indeterminacy.

Definition 7: Neutrosophic Point Graph [11]

A neutrosophic point graph $G_N$ is a graph G with finite non empty set $V_N = V_N (G)$ of p-points where at least one of the points in $V_N (G)$ is an indeterminate node, element, point or vertex.

Note here that $V_N (G) = V (G) + N$, where $V (G)$ are points or vertices of the graph $G$ and $N$ the non-empty set of points which are indeterminate nodes.

Definition 8: Neutrosophic Edge Graph [11]

Let $V (G)$ be the set of all vertices of the graph $G$. If the edge set is $E (G)$, where at least one of the edges of $G$ is an indeterminate one, then we call such graphs as neutrosophic edge graphs.

Thus, the neutrosophic vertex graph is distinctly different from the neutrosophic edge graph. They differ from each other on the edge set and the vertex set. The edge set of a neutrosophic vertex graph are all usual edges whereas only the vertex sets are indeterminate. On the contrary, the vertex set of the neutrosophic edge graph has the vertex set to be the usual set. The difference lies only in the edge set, where some of the edges are indeterminate [11]. A neutrosophic edge graph is simply referred to as a neutrosophic graph.

Definition 9: Neutrosophic Graph [11]

A neutrosophic graph is a graph in which at least one edge is an indeterminacy denoted by dotted lines.

Definition 10: Neutrosophic Directed Graph [11]

A neutrosophic directed graph is a directed graph which has at least one edge to be an indeterminacy.

Definition 11: Doubly or Strongly Neutrosophic Graph [11]

A graph $G$ is said to be a doubly or strongly neutrosophic graph if the graph has both indeterminate vertices and indeterminate edges. The indeterminate edges are denoted by dotted lines whereas the indeterminate vertices are denoted by $N_1, \ldots, N_k$.

Definition 12: Neutrosophic Vertex Graph [11]

A neutrosophic vertex graph $G_N$ is said to be neutrosophic simple if the graph has no loops and no multiple edges connecting an indeterminate vertex or two indeterminate vertices.

Definition 13: Neutrosophic Vertex [11]

Elements of $V_N (G)$, where $V_N (G)$ is a neutrosophic point graph, are called the neutrosophic vertices of $G$. The number of elements in $V_N(G)$ is $n(G) + N$, where $n(G)$ is called the order of $G$ and $N$ is the number of indeterminate nodes used in $V_N(G)$.

### B. *Degrees of truth, indeterminacy and falsehood*

An edge represents a relationship in a neutrosophic semantic net being used to represent a sentence. The vertices represent words. Further, a vertex can be a noun, or an adjective, whereas an edge can be a verb or an adverb.

The membership of a vertex in a relationship can be either true or indeterminate or false. However, it can be ambiguous as well, when this membership is either both true and indeterminate, or both false and indeterminate. For example, in the sentence “He is probably a very good person”, the membership of “He” is true to a certain degree as denoted by “very” and yet it is indeterminate to a certain degree as denoted by “probably”. Therefore, the need for all three values *t*, *i* and *f* to be associated with each word in a sentence arises. To represent these three values, a vertex membership set of 3 values {*t*, *i*, *f*} needs to be associated with each vertex. In this set the first element represents only degree of true membership, the second element represents degree of indeterminate membership, and the third element represents degree of false membership, for each vertex $V$.

### C. Degrees of positivity, neutrality and negativity

Further, neutrosophy is ambiguously used to simultaneously represent the triad concepts of truth, indeterminacy and falsehood, as well as, the triad concepts of positivity, neutrality and negativity. The term polarity is being introduced to distinguish the latter triad from the former.

Definition 14: Polarity

Polarity is defined as the term representing the polar concepts of positivity and negativity with a continuum of neutralities between the poles. It is a subset of neutrosophy.

Figure 1 shows the orientation of polarity with respect to crisp set theory, fuzzy theory and neutrosophy.

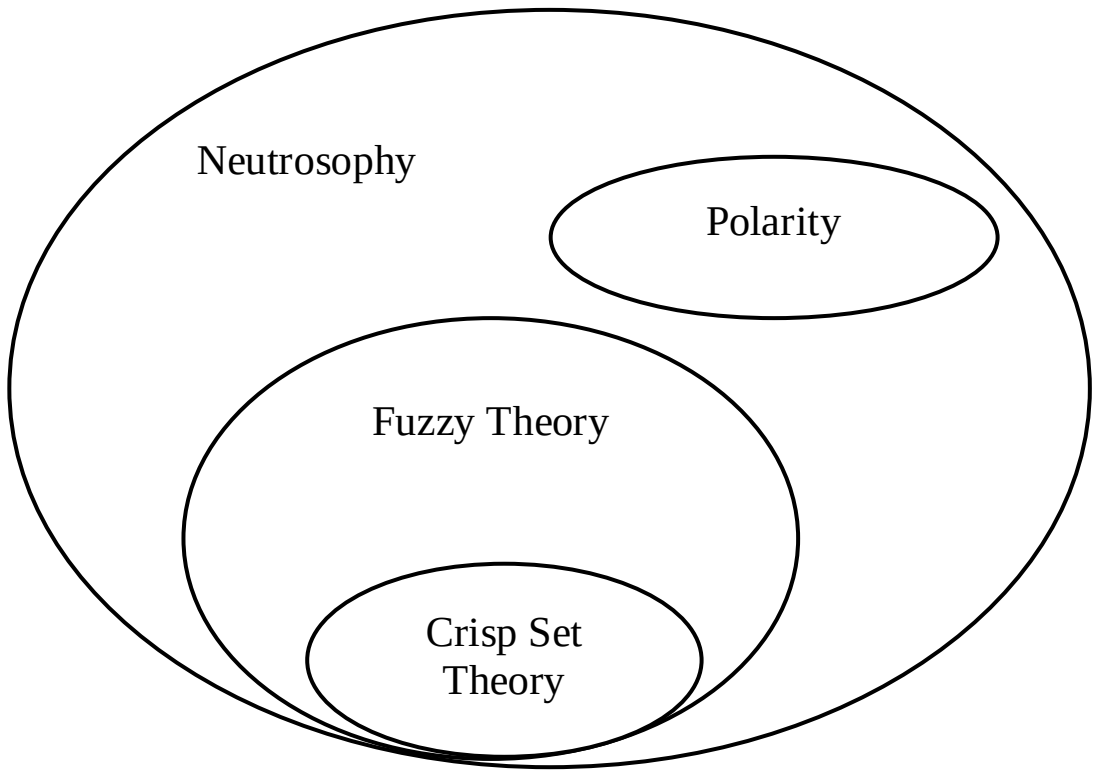

Fig. 1. Orientation of Polarity

#### 1) Degrees of positivity, neutrality and negativity of a vertex

The three notions of truth, indeterminacy and falsehood can be substituted by the notions of positivity, neutrality and negativity, respectively. That is, the values $t$, $i$ and $f$ will be replaced by $p$, $u$ and $n$, respectively, denoting the degree of positivity, the degree of neutrality and the degree of negativity. Thus, a polar vertex membership subset $M_V$ is associated with each vertex $V$. Each $M_V$ consists of 3 values ($p$, $u$, $n$). The set $M$ is the polar vertex membership set associated with a graph $G$ having a set of vertices $V$. Each vertex of $G$ is then called as a polar neutrosophic vertex.

#### 2) Degree of positivity, neutrality and negativity of an edge

Similar to the manner in which neutrosophic vertices were extended, neutrosophic edges need to be extended as well. That is, a set of three values ($p$, $u$, $n$) need to be associated with each edge.

Definition 15: Binary Neutrosophic Relation (Edges) [12]

A binary neutrosophic relation $R_N(X, Y)$ is a binary relation which assigns to each element of $X$ two or more elements of $Y$ or the indeterminate $I$. The notion of an adjacency matrix is associated with edges for denoting interconnections between them.

Definition 16: Adjacency matrix [11]

The edges $E$ of an undirected graph $G$ induce a symmetric binary relation $B$ on $V$ that is called the adjacency matrix of $G$. Specifically, for each edge $\{V_i, V_j\}$ the vertices $V_i$ and $V_j$ are said to be adjacent to one another.

Definition 17: Neutrosophic Matrix [12]

A neutrosophic matrix implies a matrix whose entries are from the set $N = [0, 1] \cup I$.

Definition 18: Fuzzy Neutrosophic Matrix [12]

A fuzzy neutrosophic matrix implies a matrix whose entries are from $N' = [0, 1] \cup \{nI \mid n \cup (0, 1)\}$.

Definition 19: Neutrosophic Adjacency Matrix [11]

Let $G$ be a neutrosophic graph. The adjacency matrix of $G$ with entries from the set ($I$, 0, 1) is called the neutrosophic adjacency matrix of the graph.

To represent these three values, a 3-dimensional neutrosophic adjacency matrix $A_{ijk}$ is necessary, in which the first element in the third dimension $A_{ij1}$ represents only $t$ values, the second element $A_{ij2}$ represents $i$ values, and the third element $A_{ij3}$ represents $f$ values.

Definition 20: Fuzzy Neutrosophic Adjacency Matrix

Let $G$ be a neutrosophic graph. The adjacency matrix of $G$ with three entries ($t$, $i$, $f$) each from the set $N' = [0, 1] \cup \{nI \mid n \in (0, 1]\}$ for each edge of G is called a fuzzy neutrosophic adjacency matrix.

Definition 21: Polar Neutrosophic Adjacency Matrix

Let $G$ be a neutrosophic graph. The extended adjacency matrix of G with three entries ($p$, $u$, $n$) each from the set $N = [0, 1] \cup I$ for each edge of $G$ is called a polar neutrosophic adjacency matrix.

Definition 22: Polar Fuzzy Neutrosophic Adjacency Matrix

Let $G$ be a neutrosophic graph. The extended adjacency matrix of $G$ with three entries ($p$, $u$, $n$) each from the set $N' = [0, 1] \cup \{nI \mid n \in (0, 1]\}$ for each edge of $G$ is called a polar fuzzy neutrosophic adjacency matrix.

## IV. Polar Fuzzy Neutrosophic Semantic Net

Definition 23: Fuzzy Neutrosophic Semantic Net (FNSN)

A FNSN is defined as a strongly neutrosophic graph $G(V, E)$ in which a fuzzy neutrosophic membership set $M^F$ is associated with $V$ and a fuzzy neutrosophic adjacency matrix $A^F$ is associated with $E$.

Definition 24: Polar Neutrosophic Semantic Net (PNSN)

A PNSN is defined as a strongly neutrosophic graph $G(V, E)$ in which a polar neutrosophic membership set $M^P$ is associated with V and a polar neutrosophic adjacency matrix $A^P$ is associated with $E$.

Definition 25: Polar Fuzzy Neutrosophic Semantic Net (PFNSN)

A PFNSN is defined as a strongly neutrosophic graph *G (V, E)* in which a polar fuzzy neutrosophic membership set $M^U$ is associated with *V* and a polar fuzzy neutrosophic adjacency matrix $A^U$ is associated with *E*. A summary of the above terminology is given in Table 1.

### A. A Polar Fuzzy Neutrosophic Semantic Net as an extended Semantic Net

The correspondence shown in Table 2, between the mathematical concepts underlying a semantic net and those underlying a PFNSN, clearly imply that a PFNSN is an extension of a semantic net.

The difference between a PFNSN and a traditional semantic net (TSN) is that a TSN cannot represent the concepts of neutrosophy and polarity. This is a serious drawback of a TSN since most of the sentences that human beings use in various situations are either neutrosophic or polar or polar neutrosophic.

### B. Illustration of the representation of a sentence using a Fuzzy Neutrosophic Semantic Net

Consider the sentence S1: “The night is rather cold and somewhat hazy but it is not raining”.

TABLE I. SUMMARY OF TERMINOLOGY

| Semantic Net type | Membership Vertex set | Adjacency Matrix |
|---|---|---|
| Fuzzy Neutrosophic Semantic Net | Fuzzy Neutrosophic Membership Vertex Set | Fuzzy Neutrosophic Adjacency Matrix |
| Polar Neutrosophic Semantic Net | Polar Neutrosophic Membership Vertex Set | Polar Neutrosophic Adjacency Matrix |
| Polar Fuzzy Neutrosophic Semantic Net | Polar Fuzzy Neutrosophic Membership Vertex Set | Polar Fuzzy Neutrosophic Adjacency Matrix |

TABLE II. CORRESPONDENCE BETWEEN THE MATHEMATICAL CONCEPTS UNDERLYING A SEMANTIC NET AND THE MATHEMATICAL CONCEPTS UNDERLYING A PFNSN

| Concepts of Linguistics | Equivalent Mathematical Concepts of a Semantic Net | Equivalent Mathematical Concepts of a Polar Fuzzy Neutrosophic Semantic Net |
|---|---|---|
| Semantic Net | Graph | Neutrosophic Graph |
| Concept or object | Node or vertex | Polar Fuzzy Neutrosophic Vertex |
| Degree of participation in a relationship | Not Represented | Degree of positivity, neutrality and negativity of the membership of a vertex in a relationship-Polar Fuzzy Neutrosophic Membership Vertex Set |
| Relationship | Binary Relation (Edges) | Binary Neutrosophic Relation (Edges) |
| Degree of Polarity | Not Represented | Degree of positivity, neutrality and negativity of the relationship between two vertices |
| Semantics of a sentence | Adjacency Matrix or Connection Matrix | Polar Fuzzy Neutrosophic Adjacency Matrix |

S1 is a moderately complex sentence which describes the weather at night. The weather is rather cold (truth *t* to a degree of 2.4), somewhat hazy (indeterminate *i* to a degree of 1.4) and not raining (false *f* that it is raining to a degree of 1.0). Thus this sentence exhibits neutrosophy (*t*, *i*, *f*) but the degrees of *t*, *i* and *f* are fuzzy. Hence the sentence is fuzzy neutrosophic and can be sufficiently represented by a FNSN. Table 3 shows the elements of S1 required for creating the FNSN for S1. Implementation of an FNSN was done in MATLAB and S1 was represented as shown in Figure 2.

TABLE III. THE ELEMENTS OF THE FNSN FOR THE SENTENCE S1.

| Input/ Output | Mathematical Concepts of a Fuzzy Neutrosophic Semantic Net | Elements of the sentence S1 |
|---|---|---|
| Inputs | Fuzzy Neutrosophic Vertices | night, cold, hazy, raining |
| | Degrees of truthhood(t), indeterminacy(i) and falsehood(f) of the membership of a vertex in a relationship- Fuzzy Neutrosophic Membership Vertex Set | $\begin{bmatrix} & t & i & f \\ night & 3.0 & 0 & 0 \\ cold & 3.0 & 0 & 0 \\ hazy & 3.0 & 0 & 0 \\ raining & 0 & 0 & 1.0 \end{bmatrix}$ |
| | Binary Neutrosophic Relation (Edges) | rather(2.4), somewhat(1.4), not(1.0) |
| | Degrees of truthhood(t), indeterminacy(i) and falsehood(f) of the relationship between two vertices | Represented using a fuzzy neutrosophic adjacency matrix as in the next row of this table. |
| | Fuzzy Neutrosophic Adjacency Matrix<br>$A_{ij1}$ represents t, $A_{ij2}$ represents i, $A_{ij3}$ represents f | $A_{ij1} = \begin{bmatrix} 0 & 2.4 & 0 & 0 \\ 0 & 0 & 0 & 0 \\ 0 & 0 & 0 & 0 \\ 0 & 0 & 0 & 0 \end{bmatrix}$ $A_{ij2} = \begin{bmatrix} 0 & 0 & 1.4 & 0 \\ 0 & 0 & 0 & 0 \\ 0 & 0 & 0 & 0 \\ 0 & 0 & 0 & 0 \end{bmatrix}$ $A_{ij3} = \begin{bmatrix} 0 & 0 & 0 & 1.0 \\ 0 & 0 & 0 & 0 \\ 0 & 0 & 0 & 0 \\ 0 & 0 & 0 & 0 \end{bmatrix}$ |
| Output | Neutrosophic Graph | See Figure 2 |

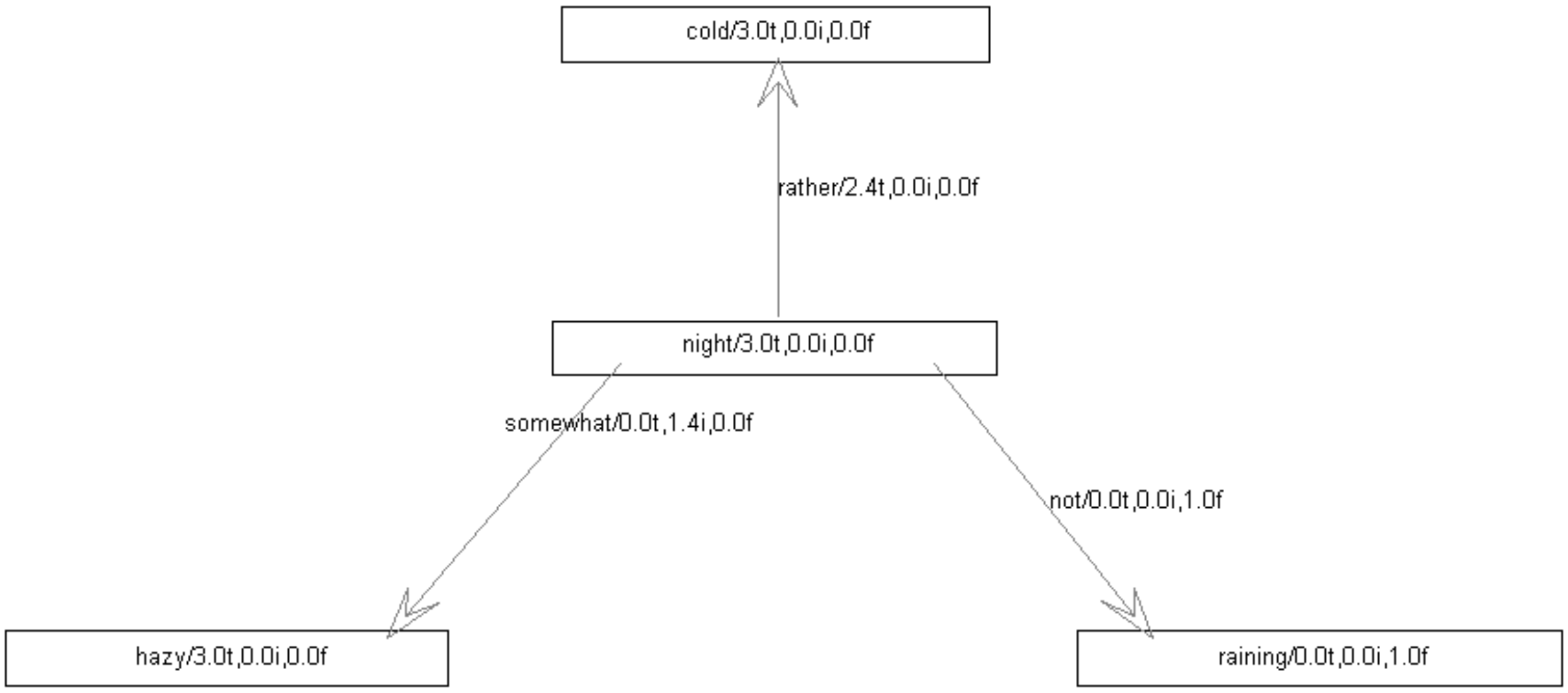

Fig. 2. Representation of the sentence S1 using MATLAB.

### *C. Illustration of the representation of a sentence using a Polar Neutrosophic Semantic Net*

Consider the sentence S2: “An atom has protons, neutrons and electrons, where protons are positive, neutrons are neutral and electrons are negative”.

S2 is a complex sentence which describes the components of an atom. These components, namely, protons, neutrons and electrons, are either completely part of an atom or completely not part of an atom, on the one hand, but are positively charged, neutral or negatively charged on the other. The charges represent the concept of polarity. But since the membership is complete in all cases, S2 does not exhibit fuzziness. Hence the sentence is polar neutrosophic.

Table 4 shows the elements of S2 required for creating the PNSN for S2. Implementation of a PNSN was done in MATLAB and S2 was represented as shown in Figure 3.

TABLE IV. THE ELEMENTS OF THE PNSN FOR THE SENTENCE S2.

| **Input/ Output** | **Mathematical Concepts of a Polar Neutrosophic Semantic Net** | **Elements of the sentence S2** |
|---|---|---|
| Inputs | Polar Neutrosophic Vertices | protons, positive, neutrons, neutral, electrons, negative, atom |
| | Positivity(p), neutrality(u) and negativity(n) of a vertex taking part in a relationship-Polar Neutrosophic Membership Vertex Set | $\begin{bmatrix} p & u & n \\ 3.0 & 0 & 0 \\ 3.0 & 0 & 0 \\ 0 & 2.0 & 0 \\ 0 & 2.0 & 0 \\ 0 & 0 & 1.0 \\ 0 & 0 & 1.0 \\ 0 & 2.0 & 0 \end{bmatrix}$ |
| | Binary Neutrosophic Relation (Edges) | are, has, are, has, are, has |
| | Positivity(p), neutrality(u) and negativity(n) of the relationship between two vertices | Represented using a polar neutrosophic adjacency matrix as in the next row of this table. |
| | Polar Neutrosophic Adjacency Matrix<br>$A_{ij1}$ represents p, $A_{ij2}$ represents u, $A_{ij3}$ represents n | $A_{ij1} = \begin{bmatrix} 0&0&0&0&0&0&0 \\ 0&0&0&0&0&0&0 \\ 0&0&0&0&0&0&0 \\ 0&0&0&0&0&0&0 \\ 0&0&0&0&0&0&0 \\ 0&0&0&0&0&0&0 \\ 0&0&0&0&0&0&0 \end{bmatrix}$ $A_{ij2} = \begin{bmatrix} 0&2.0&0&0&0&0&2.0 \\ 0&0&0&0&0&0&0 \\ 0&0&0&2.0&0&0&2.0 \\ 0&0&0&0&0&0&0 \\ 0&0&0&0&0&2.0&2.0 \\ 0&0&0&0&0&0&0 \\ 0&0&0&0&0&0&0 \end{bmatrix}$ |
| | | $A_{ij3} = \begin{bmatrix} 0 & 0 & 0 & 0 & 0 & 0 & 0 \\ 0 & 0 & 0 & 0 & 0 & 0 & 0 \\ 0 & 0 & 0 & 0 & 0 & 0 & 0 \\ 0 & 0 & 0 & 0 & 0 & 0 & 0 \\ 0 & 0 & 0 & 0 & 0 & 0 & 0 \\ 0 & 0 & 0 & 0 & 0 & 0 & 0 \\ 0 & 0 & 0 & 0 & 0 & 0 & 0 \end{bmatrix}$ |
| Output | Neutrosophic Graph | See Figure 3 |

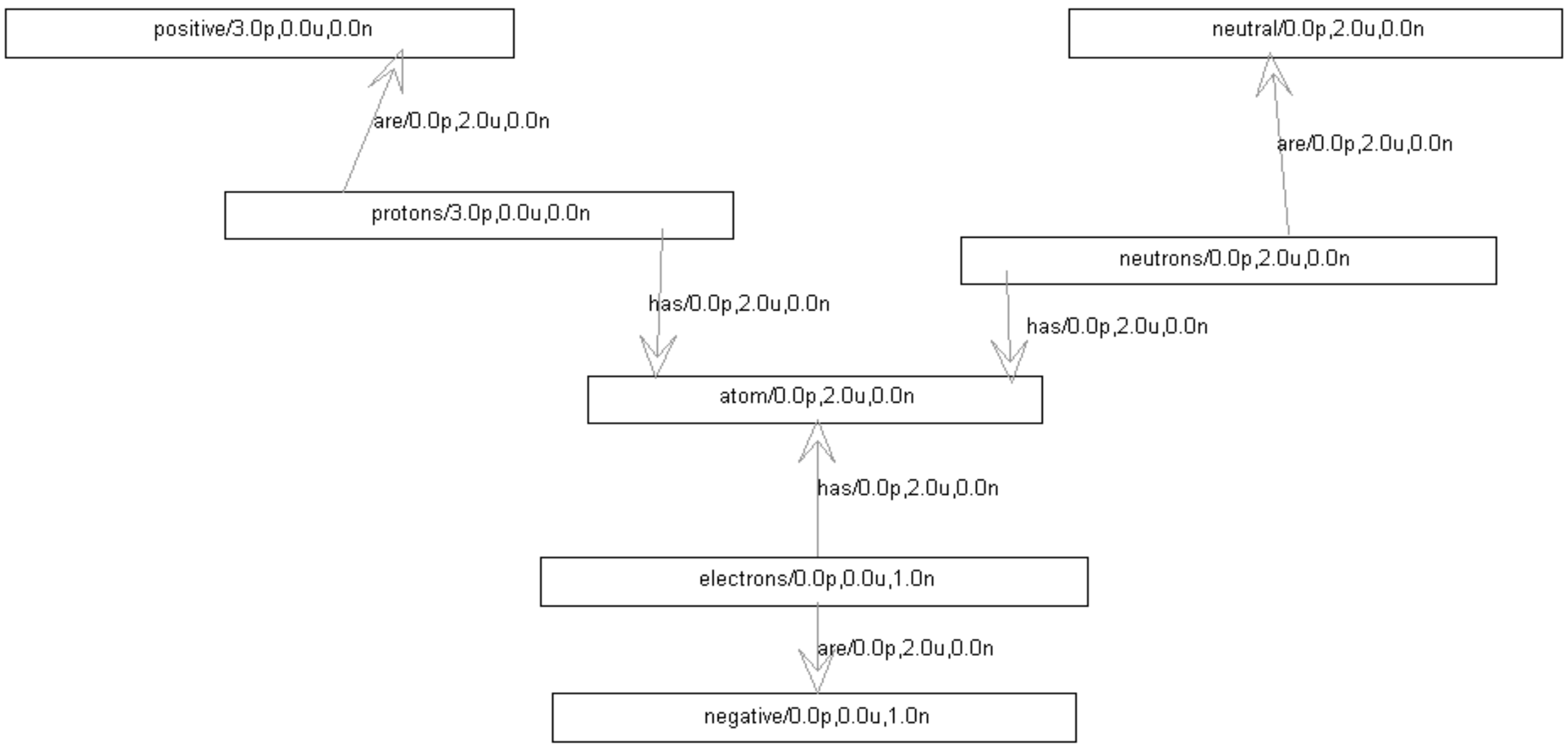

Fig. 3. Representation of the sentence S2 using MATLAB.

TABLE V. THE ELEMENTS OF THE PFNSN FOR THE SENTENCE S3.

| **Input/ Output** | **Mathematical Concepts of a Polar Fuzzy Neutrosophic Semantic Net** | **Elements of the sentence S3** |
|---|---|---|
| Inputs | Polar Fuzzy Neutrosophic Vertices | Bob is quite healthy, rather plump but slightly anaemic |
| | Degrees of positivity(p), neutrality(u) and negativity(n) of the membership of a vertex in a relationship-Polar Fuzzy Neutrosophic Membership Vertex Set | $\begin{bmatrix} & p & u & n \\ Bob & 3.0 & 0 & 0 \\ healthy & 3.0 & 0 & 0 \\ plump & 3.0 & 0 & 0 \\ anaemic & 0 & 0 & 1.0 \end{bmatrix}$ |
| | Binary Neutrosophic Relation (Edges) | quite (2.7), rather(1.4), slightly(0.3) |
| | Degrees of positivity(p), neutrality(u) and negativity(n) of the relationship between two vertices | Represented using a polar fuzzy neutrosophic adjacency matrix as in the next row of this table. |
| | Polar Fuzzy Neutrosophic Adjacency Matrix<br>$A_{ij1}$ represents p, $A_{ij2}$ represents u, $A_{ij3}$ represents n | $A_{ij1} = \begin{bmatrix} 0 & 2.7 & 0 & 0 \\ 0 & 0 & 0 & 0 \\ 0 & 0 & 0 & 0 \\ 0 & 0 & 0 & 0 \end{bmatrix}$ $A_{ij2} = \begin{bmatrix} 0 & 0 & 1.4 & 0 \\ 0 & 0 & 0 & 0 \\ 0 & 0 & 0 & 0 \\ 0 & 0 & 0 & 0 \end{bmatrix}$ $A_{ij3} = \begin{bmatrix} 0 & 0 & 0 & 0.3 \\ 0 & 0 & 0 & 0 \\ 0 & 0 & 0 & 0 \\ 0 & 0 & 0 & 0 \end{bmatrix}$ |
| Output | Neutrosophic Graph | See Figure 4 |

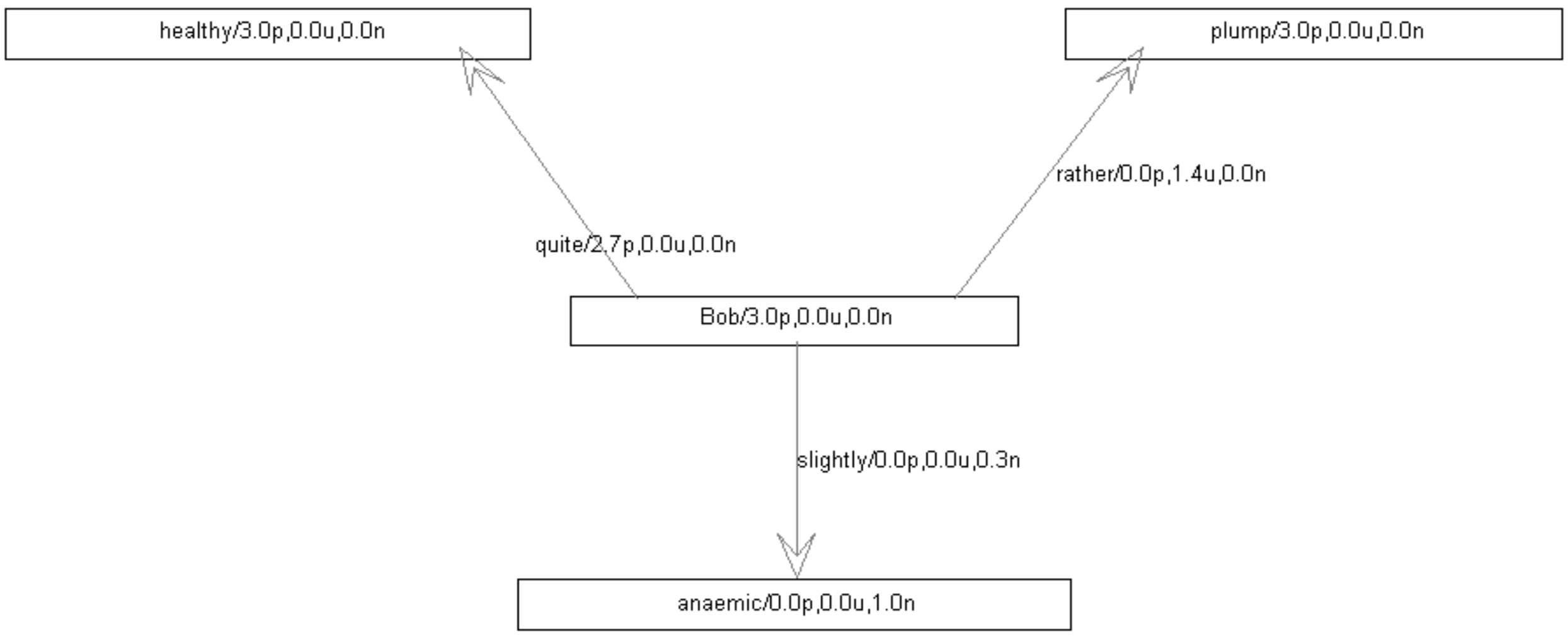

Fig. 4. Representation of the sentence S3 using MATLAB.

### D. *Illustration of the representation of a sentence using a Polar Fuzzy Neutrosophic Semantic Net*

Consider the sentence S3: "Bob is quite healthy, rather plump but slightly anaemic".

S3 is a moderately complex sentence which describes the state of Bob's health. He is quite healthy (something positive but to a degree of 2.7), rather plump (something neutral to a degree of 1.4), but slightly anaemic (something negative, to a degree of 0.3, as indicated by the conjunction "but", which is creating an opposing notion). Since the degrees of polarity are fuzzy, therefore, the sentence is polar fuzzy neutrosophic.

Table 5 shows the elements of the sentence S3 required for creating the PFNSN for S3.

Implementation of a PFNSN was done in MATLAB and the above sentence was represented as shown in Figure 4.

## V. Discussion

Each of the sentences S1, S2 and S3 have a basic structure which corresponds to a "picture" created by a human being in his mind when he creates a sentence for a given situation. Further, the structure, along with various degrees, defines the emotions with which a person utters the sentence as speech, after framing it. This lends the naturalness property to the voice of a person. The question of the applicability of fuzzy theory to speech processing has been explored in [13].

The fundamental idea underlying emotions is that they are positive, neutral or negative. A PFNSN can be directly applied for emotion representation in a machine. Further, since an intelligent response is characterized by answers that are based on an understanding of the positive and the negative aspects of the situation in which a sentence was spoken to a human being, a PFNSN can be used to generate responses which are nearer to being termed as "intelligent". A response can be considered intelligent if it takes into account positive as well as negative aspects of a situation. Aspects related to the applicability of fuzzy theory to intelligent response generation have been discussed in [14].

This can be done by decision making based on the polar selection of that vertex out of multiple vertices, which are connected to a given vertex.

## VI. Conclusion

The incorporation of fuzziness, polarity and neutrosophy into a conventional semantic net, leading to a FNSN, a PNSN and a PFNSN is a major enhancement in terms of representational capability. This enhancement holds potential to incorporate emotion representation in an otherwise emotionless robot. This representation is to the extent of emotions being represented in the form of continua for each of the positive and negative poles and neutrality. This excludes the possibility of imbibing feelings, the physical outcome of emotion generation in a living being. Further, the speech of a robot can be imbibed with naturalness. As an extension of the speech, more humane expressions can be exhibited by a robot. Thus natural language processing using these extended semantic nets can form the core of developing human-like robots.

However, there are bound to be limitations on how well these semantic nets can represent emotions and the degree to which they can be scaled. On the other hand, this enhancement holds potential towards the development of more humane robots.

## VII. Future Scope

The human mind receives inputs from the five senses. The inputs are in the form of images, video, audio, odour and touch. The inputs come together in the mind to form a picture of the current environment. This picture is the equivalent of an extended textual semantic net. These inputs are stimuli, to which the mind responds with emotions, which can be positive, neutral or negative, to qualify relationships between the inputs. Representation of these five inputs and incorporation of emotions into such a PFNSN is a promising future application area of the topic of this paper. This representation will form a part of the response subsystem of a robot.